\pdfoutput=1

\documentclass[11pt]{article}

\usepackage{emnlp2022}

\usepackage{times}
\usepackage{latexsym}

\usepackage[T1]{fontenc}

\usepackage[utf8]{inputenc}

\usepackage{microtype}

\usepackage{listings} 
\usepackage{graphicx}
\usepackage{caption}
\usepackage{subcaption} 
\usepackage{xcolor}
\definecolor{BrickRed}{rgb}{.72,0,0} 

\usepackage[utf8]{inputenc}
\usepackage[T1]{fontenc}
\usepackage{hyphenat}
\usepackage{xspace}
\usepackage{amsmath}
\usepackage{amsfonts}
\usepackage{hyperref}
\usepackage{url}
\usepackage{booktabs}
\usepackage{multirow}
\usepackage{makecell}
\usepackage{caption}
\usepackage{minibox}
\usepackage{bbm}
\usepackage{graphicx}
\usepackage{balance}
\usepackage{mathtools}
\usepackage{color}
\usepackage{marvosym}
\usepackage{ifthen}
\usepackage{textcomp}
\usepackage{enumitem}
\usepackage{verbatim}
\usepackage{algorithm}
\usepackage{numprint}
\usepackage{balance}

\usepackage{amsthm}
\theoremstyle{plain}

\newcommand{\chatoDisplayMode}[1]{#1}

\definecolor{MyRed}{rgb}{0.6,0.0,0.0} 
\definecolor{MyBlack}{rgb}{0.1,0.1,0.1} 
\newcommand{\inred}[1]{{\color{MyRed}\sf\textbf{\textsc{#1}}}}
\newcommand{\frameit}[2]{
  \begin{center}
  {\color{MyRed}
  \framebox[.9\columnwidth][l]{
    \begin{minipage}{.85\columnwidth}
    \inred{#1}: {\sf\color{MyBlack}#2}
    \end{minipage}
  }\\
  }
  \end{center}
}

\newcommand{\note}[2][]{\chatoDisplayMode{\def\@tmpsig{#1}\frameit{{\Pointinghand} Note}{#2\ifx \@tmpsig \@empty \else \mbox{ --\em #1}\fi}}}
\newcommand{\todo}[2][]{\chatoDisplayMode{\def\@tmpsig{#1}\frameit{{\Writinghand} To-do}{#2\ifx \@tmpsig \@empty \else \mbox{ --\em #1}\fi}}}

\newcommand{\abbrevStyle}[1]{#1}

\newcommand{\ie}{\abbrevStyle{i.e.}\xspace}

\newcommand{\etc}{\abbrevStyle{etc.}\xspace}

\newcommand{\Secref}[1]{Sec.~\ref{#1}}

\newcommand{\Tabref}[1]{Table~\ref{#1}}
\newcommand{\Figref}[1]{Fig.~\ref{#1}}

\newcommand{\Appref}[1]{Appendix~\ref{#1}}

\newcommand{\xhdr}[1]{\vspace{1.7mm}\noindent{{\bf #1.}}}

\newcommand{\textcite}[1]{\citeauthor{#1} \shortcite{#1}}

\newcommand{\cpt}[1]{\textsc{\MakeLowercase{#1}}}

\newcommand{\hide}[1]{}

\makeatletter
\newcommand{\iffont}[2]{\ifthenelse{\equal{\f@family}{#1}}{#2}{}}
\makeatother

  \usepackage{mathptmx}

  \DeclareSymbolFont{greek}{OML}{cmm}{m}{n}
  \DeclareMathSymbol{\alpha}{\mathalpha}{greek}{"0B}
  \DeclareMathSymbol{\beta}{\mathalpha}{greek}{"0C}
  \DeclareMathSymbol{\gamma}{\mathalpha}{greek}{"0D}
  \DeclareMathSymbol{\delta}{\mathalpha}{greek}{"0E}
  \DeclareMathSymbol{\epsilon}{\mathalpha}{greek}{"0F}
  \DeclareMathSymbol{\zeta}{\mathalpha}{greek}{"10}
  \DeclareMathSymbol{\eta}{\mathalpha}{greek}{"11}
  \DeclareMathSymbol{\theta}{\mathalpha}{greek}{"12}
  \DeclareMathSymbol{\iota}{\mathalpha}{greek}{"13}
  \DeclareMathSymbol{\kappa}{\mathalpha}{greek}{"14}
  \DeclareMathSymbol{\lambda}{\mathalpha}{greek}{"15}
  \DeclareMathSymbol{\mu}{\mathalpha}{greek}{"16}
  \DeclareMathSymbol{\nu}{\mathalpha}{greek}{"17}
  \DeclareMathSymbol{\xi}{\mathalpha}{greek}{"18}
  \DeclareMathSymbol{\pi}{\mathalpha}{greek}{"19}
  \DeclareMathSymbol{\rho}{\mathalpha}{greek}{"1A}
  \DeclareMathSymbol{\sigma}{\mathalpha}{greek}{"1B}
  \DeclareMathSymbol{\tau}{\mathalpha}{greek}{"1C}
  \DeclareMathSymbol{\upsilon}{\mathalpha}{greek}{"1D}
  \DeclareMathSymbol{\phi}{\mathalpha}{greek}{"1E}
  \DeclareMathSymbol{\chi}{\mathalpha}{greek}{"1F}
  \DeclareMathSymbol{\psi}{\mathalpha}{greek}{"20}
  \DeclareMathSymbol{\omega}{\mathalpha}{greek}{"21}
  \DeclareMathSymbol{\varepsilon}{\mathalpha}{greek}{"22}
  \DeclareMathSymbol{\vartheta}{\mathalpha}{greek}{"23}
  \DeclareMathSymbol{\varpi}{\mathalpha}{greek}{"24}
  \DeclareMathSymbol{\varrho}{\mathalpha}{greek}{"25}
  \DeclareMathSymbol{\varsigma}{\mathalpha}{greek}{"26}
  \DeclareMathSymbol{\varphi}{\mathalpha}{greek}{"27}
  \DeclareSymbolFont{otone}{OT1}{cmr}{m}{n}
  \DeclareMathSymbol{\Gamma}{\mathalpha}{otone}{0}
  \DeclareMathSymbol{\Delta}{\mathalpha}{otone}{1}
  \DeclareMathSymbol{\Theta}{\mathalpha}{otone}{2}
  \DeclareMathSymbol{\Lambda}{\mathalpha}{otone}{3}
  \DeclareMathSymbol{\Xi}{\mathalpha}{otone}{4}
  \DeclareMathSymbol{\Pi}{\mathalpha}{otone}{5}
  \DeclareMathSymbol{\Sigma}{\mathalpha}{otone}{6}
  \DeclareMathSymbol{\Upsilon}{\mathalpha}{otone}{7}
  \DeclareMathSymbol{\Phi}{\mathalpha}{otone}{8}
  \DeclareMathSymbol{\Psi}{\mathalpha}{otone}{9}
  \DeclareMathSymbol{\Omega}{\mathalpha}{otone}{10}
  \DeclareSymbolFont{syms}{OML}{cmm}{m}{it}
  \DeclareMathSymbol{\partial}{\mathord}{syms}{"40}
  \DeclareMathAlphabet{\mathbold}{OML}{cmm}{b}{it}
  \DeclareSymbolFont{largesymbols}{OMX}{cmex}{m}{n}

\usepackage{algpseudocode} 
\usepackage{placeins}


\newcommand{\dBERT}{distil\cpt{BERT}\xspace}
\newcommand{\ROBERTA}{\cpt{R}o\cpt{BERT}a\xspace}

\newcommand{\IRMLM}{iLM\xspace}
\newcommand{\ERMLM}{eLM\xspace}
\newcommand{\EnsLM}{ensLM\xspace}
\newcommand{\MTLM}{mtLM\xspace}

\newcommand{\MASK}{\cpt{MASK}\xspace}

\newcommand{\tr}{}

\title{Invariant Language Modeling}

\DeclareSymbolFont{extraup}{U}{zavm}{m}{n}
\DeclareMathSymbol{\microsoft}{\mathalpha}{extraup}{81}
\DeclareMathSymbol{\epfl}{\mathalpha}{extraup}{83}

\author{
Maxime Peyrard,$^{\epfl}$ 
Sarvjeet Singh Ghotra,$^{\microsoft}$
Martin Josifoski,$^{\epfl}$
Vidhan Agarwal,$^{\microsoft}$
\AND
Barun Patra,$^{\microsoft}$
Dean Carignan,$^{\microsoft}$
Emre K\i{}c\i{}man,$^{\microsoft}$
Saurabh Tiwary,$^{\microsoft}$
Robert West$^{\epfl}$ \\
    $^{\epfl}$EPFL \quad $^{\microsoft}$Microsoft Corporation \\
    {\{maxime.peyrard, martin.josifoski, robert.west\}@epfl.ch} \\
    {\{saghotra, vidhan.agarwal\}@microsoft.com} \\
    {\{barun.patra, dcarig, emrek, satiwary\}@microsoft.com}\\
  }

\begin{document}
\maketitle

\begin{abstract}
Large pretrained language models are critical components of modern NLP pipelines. 
Yet, they suffer from spurious correlations,
poor out-of-domain generalization,
and biases.
Inspired by recent progress in causal machine learning, in particular the invariant risk minimization (IRM) paradigm,
we propose \emph{invariant language modeling}, a framework for learning invariant representations that generalize better across multiple environments.
In particular, we adapt a game-theoretic formulation of IRM (\emph{IRM-games})
to language models, where the invariance emerges from a specific training schedule in which all the environments compete to optimize their own environment-specific loss by updating subsets of the model in a round-robin fashion.
We focus on controlled experiments to precisely demonstrate the ability of our method to (i)~remove structured noise, (ii)~ignore specific spurious correlations without affecting global performance, and (iii)~achieve better out-of-domain generalization.
These benefits come with a negligible computational overhead compared to standard training, do not require changing the local loss, and can be applied to any language model.
We believe this framework is promising to help mitigate spurious correlations and biases in language models.
\end{abstract}

\section{Introduction}
\label{sec:introduction}
\begin{figure}[t]
    \centering
    \includegraphics[width=0.85\columnwidth]{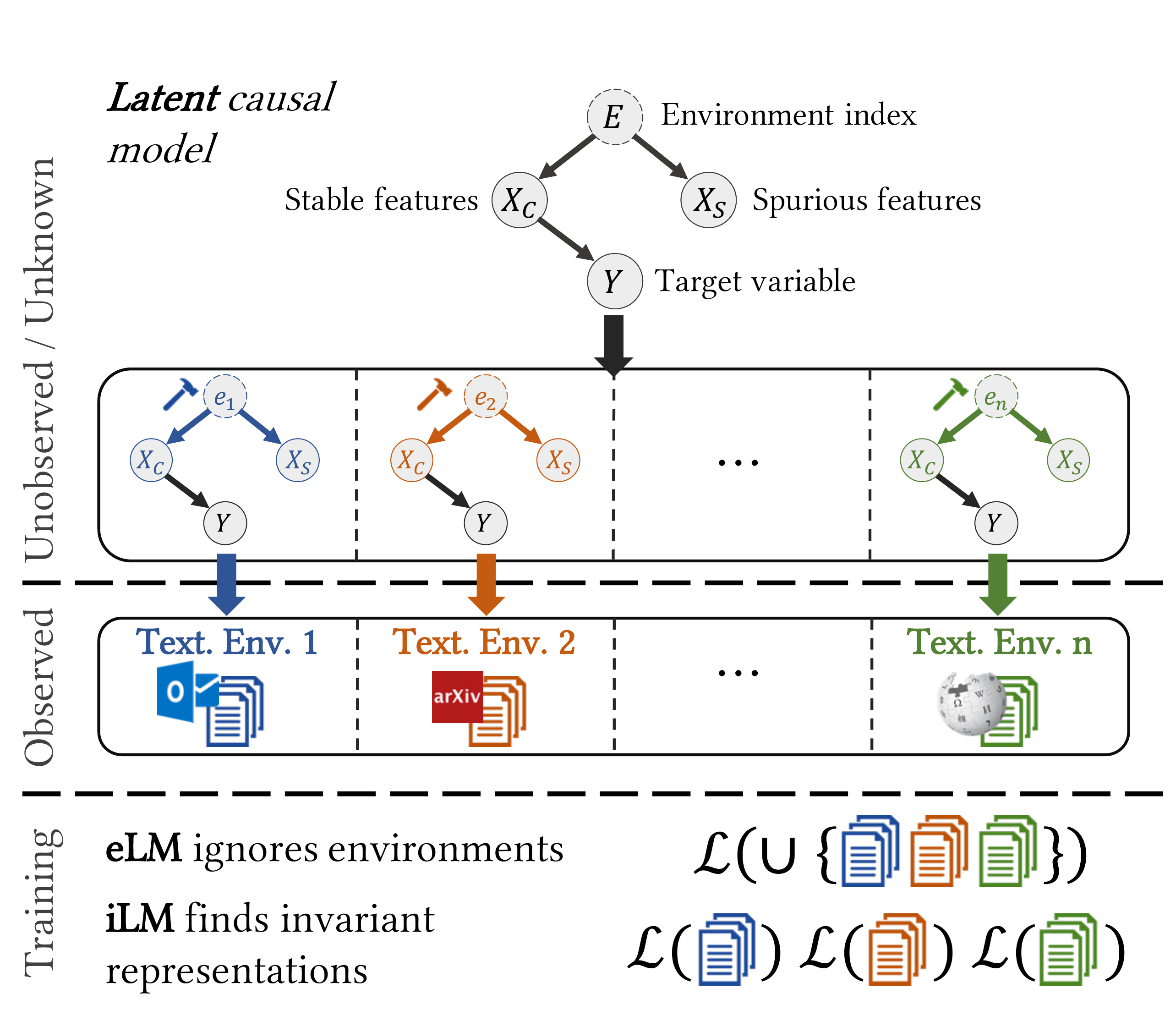}
    \caption{\textbf{High-level overview using a simplified causal structure.} 
    The distinction between environments makes it possible to separate spurious from stable features. Indeed, the relationship between the target variable $Y$ and the stable features $X_C$ is invariant across environments: $\mathbb{E}[Y|X_C, E] = \mathbb{E}[Y|X_C]$. However, the correlation between $Y$ and $X_S$ is spurious and does not generalize across environments: $\mathbb{E}[Y|X_S, E=e] \neq \mathbb{E}[Y|X_S, E=e']$ for $e \neq e'$. Language models trained with the standard ERM, denoted as \ERMLM in this work, exploit all correlations available during training and aim to learn $\mathbb{E}[Y|X_C, X_S]$.  Our proposed invariant language models, denoted as \IRMLM, focus on invariant features and aim to learn $\mathbb{E}[Y|X_C]$. 
    In language modeling, $Y$ could represent the missing-word prediction task.
    }
    \label{fig:figure_1}
\end{figure}

While modern pretrained transformer models have led to
dramatic progress on many NLP tasks, important limitations remain.
In particular, pretrained language models suffer from poor generalization, even under small perturbations of the input distribution \cite{moradi2021evaluating}. Indeed, these models encode \cite{moradi2021evaluating} and exploit \cite{tu-etal-2020-empirical, niven-kao-2019-probing} spurious correlations, i.e., correlations that do not generalize across data distributions. 
Since language models are trained on large unverified corpora, they also suffer from biases \cite{nadeem-etal-2021-stereoset,bordia-bowman-2019-identifying}.
Here the term ``biases'' refers to correlations that may or may not be spurious according to the available textual data distributions, but are nevertheless undesired.
Existing techniques aiming to remove spuriousness or biases involve computationally expensive domain alignment \cite{adv_inv_features, liu2020causal_semantic_rep, zhao2020entropy_reg}, domain transfer \cite{balaji2018metareg}, or the addition of penalty terms to the loss targeted at specific undesired correlations \cite{qian-etal-2019-reducing,zhao-etal-2018-learning}.
Alternatively, data preprocessing \cite{zhao-etal-2017-men, zhou2021feature_noise} or manipulation such as counterfacual data-augmentation \cite{cda} can yield datasets where undesired correlations are less present. 
Pretraining with larger and more diverse datasets can also help \cite{tu-etal-2020-empirical, brown2020language}.

However, recent works on the theory of causality \cite{pearl, abs-1911-10500} argue that removal of spurious correlations requires altogether different learning and training paradigms going beyond purely statistical learning.
Indeed, generalization, spuriousness, and biases are all better understood in the language of causality \cite{pearl}. 
Intuitively, causal relationships are the ones expected to be stable \cite{Scholkopfetal21, Peters2017} and generalizable \cite{icc}.
When the causal graph underlying the data generation mechanism is known, there exist causal identification algorithms to distinguish \emph{desired} from \emph{undesired} correlations \cite{JMLR:v9:shpitser08a}.
However, for complex tasks of interest, the underlying causal model is not known. Language modeling is one of these tasks, where it is unclear what would even be the relevant random variables constituting the causal model.  

Therefore, causal identification from the causal graph seems out-of-reach for language modeling. Similarly, removing undesired correlations one by one is impractical due to the sheer amount of possible correlations to consider. 
In this work, we propose to benefit from recent progress in causal machine learning to offer a new and more flexible lever for dealing with spuriousness and biases.
We take inspiration from the \emph{invariance principle,} which states that only relationships invariant across training \emph{environments} should be learned \cite{icc}. Under specific assumptions, the invariant representation would then only encode the causal relationships relevant to the task and should thus generalize. 
Environments correspond to different views of the learning task, i.e., different data distributions. 
The invariance principle is illustrated by \Figref{fig:figure_1} with a simplified causal model as an example. $E$ represents environment indices, $Y$ is the target variable, $X_C$ are the \emph{causal features}, such that $\mathbb{E}[Y|X_C]$ is stable across environments ($\mathbb{E}[Y|X_C, E] = \mathbb{E}[Y|X_C] $), and $X_S$ are the spurious features, not generalizing across environments ($\mathbb{E}[Y|X_S, E=e] \neq \mathbb{E}[Y|X_S, E=e']$ for $e \neq e'$). 
Language models trained with standard empirical risk minimization (ERM), denoted as \ERMLM in this work, exploit all correlations available during training and aim to learn $\mathbb{E}[Y|X_C, X_S]$.  Our proposed invariant language models, denoted as \IRMLM, focus on invariant features and aim to learn $\mathbb{E}[Y|X_C]$.
In practice, since the causal model is unknown, it is the choice of environments that defines what correlations are spurious. 
Invariant learning with appropriate choices of environments is the lever we propose to employ to more flexibly deal with spuriousness and biases.

A practical formulation of the invariance principle was proposed by \newcite{IRM}. They introduced \emph{invariant risk minimization} (IRM), an alternative to ERM as a training objective enforcing the learning of invariant representations. \newcite{IRMgames} later improved the training procedure to solve the IRM objective with a method called \textit{IRM-games}. Unlike previous methods for removing biases and spurious correlations, IRM-games does not modify the loss with a regularization term and does not compute domain alignment (or matching) statistics. The invariance benefits come from the specific training schedule where environments compete to optimize their own environment-specific loss by updating subsets of the model in a round-robin fashion. 

We argue that the IRM paradigm, and IRM-games specifically, is well-suited to improve NLP systems. Textual data naturally comes from different environments, e.g., encyclopedic texts, social media posts, news articles, \etc Moreover, not knowing the causal mechanisms behind language generation within these environments is not a blocker, as the relevant variables can now remain latent. 
By adapting IRM-games to language modeling, we introduce \textit{invariant language modeling (\IRMLM),} where the training of existing pretrained models is continued to enforce invariant representations, using a simple and efficient modification of the training process. We then investigate the ability of \IRMLM to deal with undesired correlations in a series of controlled experiments, answering our core \textbf{research question:} {\em Does the invariance principle give rise to a practical strategy to deal with spurious correlations in language models?}

\xhdr{Contributions}
(i)~We introduce a new training paradigm (\IRMLM) for language models based on the invariance principle (\Secref{sec:models}). Thanks to the use of the IRM-games training schedule (see \Secref{sec:background}), our \IRMLM framework results in negligible computational overhead compared to standard ERM training, does not require changing the local loss, and is agnostic to the language model architecture.
(ii)~In a series of controlled experiments (\Secref{sec:experiments}), we demonstrate the ability of \IRMLM to remove structured noise (\Secref{ssec:struct_noise}), ignore specific spurious correlations without affecting global performance (\Secref{ssec:controlled_corr}), and achieve better out-of-domain generalization (\Secref{ssec:ood}).
(iii)~We discuss our contributions in relation to previous work (\Secref{sec:discussion}).
(iv)~Finally, we release Huggingface\hyp compatible code for training \IRMLM using existing language model checkpoints \cite{wolf-etal-2020-transformers}:
\url{https://github.com/epfl-dlab/invariant-language-models}

\section{Background}
\label{sec:background}

\subsection{Invariance Across Environments (IaE)}
Recent works on the theory of causality \cite{pearl, abs-1911-10500} have argued that out-of-distribution generalization and removal of spurious correlations require going beyond purely statistical learning.
This is motivated by the intuition that causal relationships are the ones that are expected to be robust and generalizable \cite{icc}. 
In causal machine learning, these ideas crystallized in the \emph{invariance principle} which states that only relationships invariant across training environments should be learned \cite{icc, inv_feature_rep}. In this paradigm, different environments correspond to data collected in different setups, i.e., different data distributions \cite{pearl}.
\textbf{For NLP}, spurious correlations and lack of out-of-distribution generalization are particularly well-documented and important problems \cite{moradi2021evaluating, tu-etal-2020-empirical, niven-kao-2019-probing}. Fortunately, separations between environments naturally emerge in textual data: encyclopedic, news, social media, movie subtitles, \etc 
This separation makes invariance-based approaches particularly well-suited for NLP.

\subsection{Invariant Risk Minimization (IRM)}
While the invariance principle is a general and powerful idea, works based on this principle often require knowing which random variables are part of the causal model \cite{adv_inv_features, icc}.
\citet{IRM} introduced \emph{invariant risk minimization} (IRM), an alternative to empirical risk minimization (ERM), and a practical training objective \emph{enforcing invariance in the learned latent representation}. 
IRM also builds on the idea that the training data comes from different environments
$e \in \mathcal{E}_{\tr}$. Each environment $e \in \mathcal{E}_{\tr}$ induces i.i.d.\ samples $D^e$ from a distribution $P(X^e, Y^e)$. The goal, then, is to use these multiple datasets to learn a predictor $Y \approx f(X)$, which performs well across the set of all environments $\mathcal{E}^*$, only part of which were seen during training:
$\mathcal{E}_{\tr} \subset \mathcal{E}^*$.
This is accomplished by decomposing $f$ into a feature representation $\phi$ and a classifier $w$, as $f = w \circ \phi$, where $\circ$ denotes function composition.
The feature representation $\phi$ elicits an invariant representation of the data if the same classifier $w$ is simultaneously optimal for all environments $e \in \mathcal{E}_{\tr}$. 
Intuitively, $\phi$ learns a representation that is invariant with respect to the environments if its representation is \emph{equally useful} for all environments.
\textbf{For NLP}, we propose to use the main body of a language model as the invariant feature learner $\phi$. When trained on a language modeling task, $w$ will be the language modeling heads. Then, $Y$ is the masked
word and $X$ the context.

\subsection{IRM-Games}
IRM is a challenging bi-level optimization originally solved \cite{IRM} by setting the invariance criteria as a regularizer.
Later, \citet{IRMgames}
improved the training procedure by using a game-theoretic perspective in which each environment $e$ is tied to its own classifier $w^e$. A global classifier $w$ is then defined as the ensemble of all environment-specific classifiers:
$
w = \frac{1}{|\mathcal{E}_{\tr}|} \sum\limits_{e \in \mathcal{E}_{\tr}} w^e
$ (where the predictions, not the weights, are averaged).
Then, environments take turns making a stochastic gradient update to minimize their own local empirical risk, 
by \textit{updating only the weights of their own classifier} $w^e$, while the shared $\phi$ is updated periodically. For more details see the algorithm called V-IRM in the original paper. \citet{IRMgames} showed that the equilibrium of this game is a solution to the IRM objective, \ie, the resulting $\phi$ learns invariant features.
\textbf{For NLP}, we argue that IRM-games is a particularly meaningful candidate to adapt to language modeling because it requires little structural modifications.

\subsection{Why Invariance Is Needed for NLP}
Textual data is particularly subject to distribution shifts and out-of-domain distributions as texts naturally come from different environments. This creates a highly non-i.i.d.\ setting with problems of generalizability and spurious correlations. The curse becomes a blessing when moving to invariance-based ideas, as having diverse and naturally emerging environments is the necessary starting point of algorithms like IRM-games.

As a simple example, consider gender bias in pretrained language models. When the model is queried with $q=$ ``\MASK is the best doctor'', it feeds $q$ into its main body $\phi$, from which a language modeling head $w$ outputs softmax scores $w \circ \phi(q)$. Despite the context $q$ containing no gender information, existing models score the pronoun \emph{he} much higher than \emph{she}. The problem comes from the presence of spurious correlations, where the context, here the word \textit{doctor}, is correlated with \emph{he}.
In an invariance-based approach, the training data comes from different environments. Suppose there is an environment $e$ where the data is not gender-biased, \ie, there is no correlation between the latent representation $\phi(q)$ and \emph{he}. It is thus not stable across environments (not invariant) and will not be learned.
Now, consider the slightly different query $q'=$ ``\MASK is the best doctor, she is great!''.
Here, the context $\phi(q')$ contains gender information. In all environments, the pronoun \emph{she} should be preferred. This association arises not from a spurious correlation in data but from a commonsense, almost grammatical, constraint. Therefore, this correlation is invariant and will be learned by invariance-based approaches.

This exemplifies the potential benefits of invariance-based approaches and illustrates the importance of choosing environment splits appropriately. One should not expect any arbitrary split of environments to {magically} yield generalization benefits. However, the choice of environments within the invariance-based learning framework provides a flexible new lever to inject (i) inductive biases, (ii)~knowledge about the data generation mechanism, and (iii) desirable stable properties (like removing gender bias).

\section{Model}
\label{sec:models}

We introduce a way to train language models inspired by the IRM-games setup.
This involves distinguishing the shared invariant feature extractor $\phi$ from the environment\hyp specific $w_e$'s.
With modern language model architectures, a natural choice emerges: $\phi$ as the main body of the encoder, and $w_e$ as the language modeling head that outputs the logits after the last layer.

Formally, suppose we have $n$ environments consisting of data $\{(X^e, Y^e)\}_{e=1, \dots, n}$. For a batch $(x_i, y_i) \sim P(X^i, Y^i)$ from environment $i$, the model output is formed using an ensemble of $n$ language modeling heads $\{w_{e}\}_{e=1, \dots, n}$ on top of the transformer encoder:
$
   \hat{y} = \operatorname{softmax}
  \left(\frac{1}{n} \sum\limits_{e=1}^n w_{e} \circ \phi(x_i) \right).
$
Then, a (masked) language modeling loss $\mathcal{L}$ is computed on the model output $\hat{y}$.
Note that it is the predictions of the $n$ heads that are averaged not the weights or the gradients.
No head gets to predict alone; the $n$ heads always predict together as an ensemble. The heads are subject to competitive gradient updates in a round-robin fashion as described below, which in turn creates the conditions that enforce the invariance.

\xhdr{Training}
The training of \IRMLM follows the pseudo-code described in Alg.~\ref{alg:irmlm}, where environments take turns to send a batch of data and update $\phi$ and their associated head. An illustration is provided in \Appref{app:illustration}.
Each head periodically gets an opportunity to pull the global ensemble classifier $w$ and the feature learner $\phi$ towards fitting the distribution of its associated environment. Intuitively, since each head gets the same amount of updates, the game converges to a global classifier that is simultaneously optimal for each environment, as demonstrated by \citet{IRMgames}.
While the V-IRM algorithm of \newcite{IRMgames} only updates $\phi$ periodically, we found it more stable to update it together with every head update.

\begin{algorithm}
	\caption{\IRMLM training} 
	\label{alg:irmlm}
	\begin{algorithmic}[1]
	    \State Initialize$(\phi, \{w_e\}_{e \in \mathcal{E}})$
		\For {iteration $\in \{1,2,\ldots, \frac{N_{steps}}{|\mathcal{E}|}\}$}
			\For {environment $\textcolor{BrickRed}{e} \in \mathcal{E}$}
				\State $\textcolor{BrickRed}{(x_i, y_i)} \leftarrow$ GetBatchFromEnv$(e)$ 
				\State CompetitiveUpdate$(\textcolor{BrickRed}{x_i}, \textcolor{BrickRed}{y_i}, \phi,  \{w_e\}_{e \in \mathcal{E}})$
		    \EndFor	
		\EndFor
		\Function{CompetitiveUpdate$(\textcolor{BrickRed}{x_i}, \textcolor{BrickRed}{y_i}, \phi, \{w_e\})$}{}
		  
	        \State $L = \mathcal{L}\left(\operatorname{softmax} 
  \left(\frac{1}{n} \sum\limits_{e=1}^n w_{e} \circ \phi(\textcolor{BrickRed}{x_i})\right), \textcolor{BrickRed}{y_i}\right)$
            \State GradientUpdate$(L, \phi, \textcolor{BrickRed}{w_i})$
        \EndFunction
	\end{algorithmic} 
\end{algorithm}

An advantage of this implementation is that invariance is obtained with few modifications to language models. Such simplicity arises from our leveraging of IRM-games, where invariance comes from the training schedules and ensembling of classifiers. 
Furthermore, we implement two baselines that appear similar but do not enjoy the same theoretical properties: \MTLM and \EnsLM.
The multitask baseline \cite{liu-etal-2019-multi}, \MTLM, also uses data split into environments with one head per environment and each environment being seen as a different task. 
The ensemble baseline \cite{NEURIPS2018_94ef7214}, \EnsLM, has a similar architecture as \IRMLM, ensembling $n$ heads for predictions but always updating every head with every batch. The ensemble baseline has the same forward pass as \IRMLM but does not perform the \emph{competitive gradient update}.
These baselines serve as ablations of \IRMLM to demonstrate the importance of splitting the data into environments, ensembling the heads, and using the competitive gradient update.

\section{Experiments}
\label{sec:experiments}

\begin{table}
\centering
\setlength{\tabcolsep}{3pt}
\begin{tabular}{@{}l|cc@{}}
\midrule
            & \dBERT{}  & \ROBERTA{} \\ 
\midrule
\ERMLM         & 4.71{\scriptsize± .04} & 3.93 {\scriptsize± .06} \\
\MTLM         & 4.65{\scriptsize± .05} & 3.74 {\scriptsize± .05} \\
\EnsLM         & 4.66{\scriptsize± .03} & 3.79 {\scriptsize± .02} \\
\IRMLM         & \textbf{4.43{\scriptsize± .03}} & \textbf{3.66{\scriptsize± .04}} \\
\bottomrule
\end{tabular}
\caption{\textbf{Robustness to noise.} Average perplexity over hyper-parameters (lower is better).
The differences between \IRMLM{} and the others are statistically significant (paired $t$-test, $p<10^{-7}$).}
\label{tab:struct_noise}
\end{table}

Invariance training comes with the promise of robustness and generalization \cite{icc, inv_feature_rep, IRMgames}.  
In the following series of experiments, we test whether our proposed architecture for language modeling can provide such benefits.
Since our approach is agnostic to the language model, we focus on two small LMs used heavily in practice: \dBERT{}~\cite{dbert} and \ROBERTA{}~\cite{roberta}.
In this work, we do not aim to engineer the best possible LM but rather precisely test \IRMLM in controlled setups by crafting environments whose difference is known, from which we know the expected behavior.
We describe three experiments: robustness to noise, bias removal, and out-of-domain generalization.

Throughout the experiments, we report estimated uncertainties with $95$\% confidence intervals. We repeat experiments for varying hyper-parameters and different random seeds (see \Appref{app:details}).

\subsection{Robustness to Noise}
\label{ssec:struct_noise}

In this experiment, we test robustness in a controlled setup. We craft two environments: Env-A made of clean Wikipedia articles and Env-B made of full HTML pages of Wikipedia articles. We use 120K articles split equally into the two environments (see \Appref{app:struct_noise} for data details).
Then, we continue the training with the masked language modeling (MLM) loss from existing checkpoints for each of \IRMLM, \ERMLM, \MTLM, and \EnsLM with these two environments and evaluate the MLM perplexity on a held-out dataset of clean Wikipedia articles (25K held-out sentences).
Intuitively, \ERMLM should try to fit the HTML part of the training data and thus be more surprised by the clean Wikipedia articles during the test set. However, \IRMLM should learn to ignore the HTML because it does not generalize from Env-B to Env-A.

\xhdr{Results}
The results averaged over 16 hyper-parameters choices are reported in
\Tabref{tab:struct_noise}. See \Appref{app:struct_noise} for hyper-parameters considered.
For reference, the perplexities on the same test set of off-the-shelf pretrained \dBERT and \ROBERTA{} are, respectively, $14.43$ and $6.71$.
We observe that \IRMLM systematically has a significantly better test perplexity. 
Also, \EnsLM{} and \MTLM perform significantly better than \ERMLM but significantly worse than \IRMLM. This indicates that splitting data in $n$ environments and ensembling $n$ heads gives some robustness benefits. The full benefit comes when further combined with the training schedule of \IRMLM. We come back to this discussion in \Secref{ssec:ablation}.

To compare architectures over the test set with different hyper-parameters, base transformers, and random seeds, we also performed paired aggregation comparison based on the Bradley--Terry model, following the recommendations of \citet{peyrard-etal-2021-better}. The \textit{Pairformance} tool\footnote{\url{https://github.com/epfl-dlab/pairformance}} measures the probability that \IRMLM{} beats \ERMLM{} when hyper-parameters are matched. We obtain that \IRMLM{} significantly beats \ERMLM{} with $.98$ estimated probability. Similarly, \IRMLM beats \EnsLM with $.89$ estimated probability and \MTLM with $.92$ estimated probability. In these experiments, paired comparisons are particularly important because varying hyper-parameters result in large variations of perplexity, such that blindly averaging can amplify the variance and hide the structure of model performance.

\subsection{Bias Removal}
\label{ssec:controlled_corr}

In this experiment, we test the capacity to remove one precise and known correlation by crafting two environments differing only in this specific correlation. 
We use binary gendered terms and create two environments where the gendered terms are used differently.\footnote{We recognize the non-binary nature of gender as well as the many ethical principles in the design, evaluation, and reporting of results in studying gender as a variable in NLP \cite{larson-2017-gender}. Because \IRMLM is not limited to training only with two environments, this architecture can also support more general bias removal goals.} 
We follow the standard setup of Counterfactual Data Augmentation (CDA) \cite{cda}: we take a textual data source with known gender bias, in this case, Wikitext-2 \cite{MerityXBS16}.
A fraction $p$ of the data goes into Env-A, the rest ($1-p$) goes into Env-B. Env-A remains untouched and preserves all the properties of the original data source, whereas Env-B is intervened upon by inverting all gendered terms based on a dictionary provided by previous work \cite{bordia-bowman-2019-identifying}. 
When $p=1-p=0.5$ and the language model is finetuned with \ERMLM, this setup matches the CDA method \cite{cda} used to mitigate gender bias in NLP. Intuitively, \IRMLM should learn to ignore gender-based correlations no matter what the fraction $p$. However, \ERMLM is only expected to ignore them when $p=1-p=0.5$, i.e., the two environments precisely balance each other.

\begin{figure}
    \centering
    \includegraphics[width=0.85\linewidth]{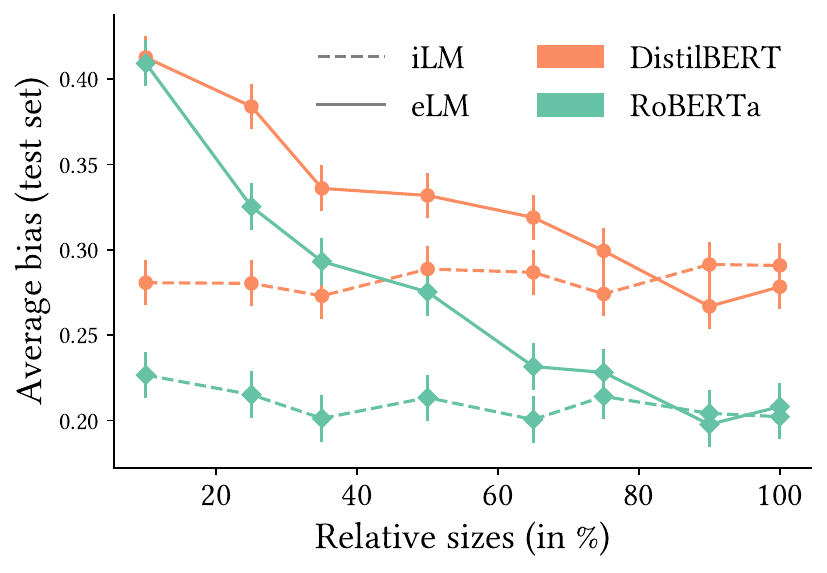}
    \caption{\textbf{Bias removal.} The x-axis represents the relative size ($x = \frac{1-p}{p}$ in percentages) between the modified environment and the unmodified one and the y-axis is the average bias for both \IRMLM and \ERMLM. Note that, according to Pairformance, $\mathbb{P}(\IRMLM \text{ beats } \ERMLM) > 0.95$ when the relative size is $< 80\%$, and that \ERMLM and \IRMLM become indistinguishable for relative sizes $>80\%$. Due to space, we report the results obtained by \EnsLM and \MTLM in \Appref{app:gb} which also shows that they perform in between \IRMLM and \ERMLM.
    }
    \label{fig:controlled_correlation}
\end{figure}

\xhdr{Experimental setup}
To measure whether the correlation has been successfully removed, we
(i) take all gendered terms in the test set, 
(ii) replace them by the \MASK token, 
(iii) use trained models to predict the missing term,
(iv) look in the softmax for the scores received by the terms of the target gendered pair. We note $s_f$ and $s_m$ the score assigned to the female and male terms in the softmax.
Finally, (v) we compute an entropy-based bias measure:
    $B_{H} = H_2\left(\frac{1}{2}\right) - H_2\left(\frac{s_f}{s_f + s_m}\right),$
where $H_2$ is the binary entropy (note that $H_2\left(\frac{1}{2}\right) = 1$). $B_{H}$ measures the extent to which a softmax has a preference for the male or female term in a gendered pair of terms. For example, in the sentence ``\MASK is the best doctor'' we look at the softmax score of the gendered-pair [\textit{he}, \textit{she}]. 
If a model has learned to ignore gender-based correlation, the entropy should be high (entropy-bias low), not favoring one gendered term over the other. We remove sentences with several gendered terms from the test set to avoid penalizing models for preferring a gender when the context contains gender information.

We ran the experiments for varying values of $p$ averaging across different hyper-parameters, and report the results in \Figref{fig:controlled_correlation} for \IRMLM and \ERMLM. The results for \EnsLM and \MTLM are reported in \Appref{app:gb}.
See \Appref{app:gb} for hyper-parameters considered.
For reference, the entropy bias of \dBERT and \ROBERTA before training are, respectively, $0.39$ and $0.46$.

\xhdr{Analysis}
Compared to off-the-shelf models, both \ERMLM and \IRMLM largely decrease the average entropy bias in the balanced setup but only \IRMLM succeeds in the unbalanced setup.
In the balanced setup (relative sizes close to $100\%$), \ERMLM and \IRMLM perform within each other's confidence intervals. 
However, in the unbalanced setup, \IRMLM largely outperforms \ERMLM. We note that, according to Pairformance, the probability that \IRMLM beats \ERMLM for any given hyper-parameter configuration is $>0.9$ for both \dBERT and \ROBERTA when the relative sizes is below $80\%$. As desired \IRMLM is not affected by the relative size of the environments.
These results confirm the hypothesis, that bias removal needs a precisely balanced dataset for \ERMLM \cite{cda}, while it does not matter for \IRMLM.
Furthermore, this entropy bias reduction does not happen at the cost of worst general perplexities (see \Appref{app:gb}). 
These findings are significant for the field of bias removal, as \IRMLM offers a practical and efficient way of removing biases. It is now not necessary to carefully counter-balance the bias in the augmented data. In \Figref{fig:controlled_correlation}, we see that already at $10\%$ of relative size, \IRMLM{} performs as well as existing approaches ($100\%$ relative size + \ERMLM{}).

\subsection{Out-of-Domain Generalization}
\label{ssec:ood}

In this experiment, we venture beyond controlled environments and test out-of-domain generalization with naturally occurring environments. 
We use \emph{thePile} dataset \citep{gao2020pile} which contains 20 very diverse textual domains: OpenSubtitles, ArXiv papers, News, GitHub comments, \etc

\xhdr{Experimental setup}
We randomly sample $11$ domains from thePile for training, the remaining $9$ domains are used for testing language models out-of-domain. Once the models are trained, using domains as environments, we evaluate their perplexity in-domain (InD) using held-out data from the training environments and OoD using data from unseen environments.
See \Appref{app:ood} for details regarding training domains and hyper-parameters. Furthermore, the trained models are evaluated on the GLUE benchmark. Indeed, models trained with \IRMLM can be used downstream exactly as if they were trained with \ERMLM. We report aggregated results in \Tabref{tab:eval-systems}.
The results also show significant improvement of \IRMLM over other architecture across the board. In particular, \IRMLM is beneficial for both in-domain (InD) and out-of-domain (OoD) evaluation.

\begin{table}[t]
\centering
\resizebox{0.85\columnwidth}{!}{
\setlength{\tabcolsep}{3pt}
\begin{tabular}{@{}lcccc@{}}
\toprule
& InD-LM$\downarrow$ & OoD-LM$\downarrow$ & GLUE$\uparrow$ \\
\midrule
\midrule
\multicolumn{3}{l}{\textbf{\dBERT}} \\ 

\hspace{4mm} \ERMLM  & 26.02{\scriptsize± 0.35} & 31.52{\scriptsize± 0.20} & 72.12  \\
\hspace{4mm} \EnsLM  & 22.31{\scriptsize± 0.56} & 32.80{\scriptsize± 0.23} & 72.34  \\
\hspace{4mm} \MTLM  & 22.73{\scriptsize± 0.29} & 31.16{\scriptsize± 0.44} & 72.22  \\
\hspace{4mm} \IRMLM  & \textbf{20.25}\textsuperscript{*}{\scriptsize± 0.52} & \textbf{30.32}\textsuperscript{*}{\scriptsize± 0.43} & \textbf{72.45} \\

\midrule
\multicolumn{3}{l}{\textbf{\ROBERTA}} \\

\hspace{4mm} \ERMLM  & 14.55{\scriptsize± 0.21} & 17.72{\scriptsize± 0.25} & 76.89 \\
\hspace{4mm} \EnsLM  & 12.40{\scriptsize± 0.34} & 17.68{\scriptsize± 0.22} & 77.49 \\
\hspace{4mm} \MTLM  & 12.56{\scriptsize± 0.33} & 17.43{\scriptsize± 0.23} & 76.55 \\
\hspace{4mm} \IRMLM  & \textbf{11.88}\textsuperscript{*}{\scriptsize± 0.28} & \textbf{16.97}\textsuperscript{*}{\scriptsize± 0.19} & \textbf{78.54}\textsuperscript{*} \\

\bottomrule                            
\end{tabular}
}
\caption{\textmd{\textbf{ThePile environment experiments.} The first column is for language modeling evaluation in-domain (perplexity, lower is better), the second column is for language modeling evaluation  out-of-domain (perplexity, lower is better), and the last column is for GLUE tasks averaged (higher is better). We mark with \textsuperscript{*} the cases where \IRMLM is statistically significantly better than other architectures (paired $t$-test).
}}
\label{tab:eval-systems}
\end{table}

\subsection{Ablation}
\label{ssec:ablation}
The \ERMLM, \MTLM, and \EnsLM architectures serve as ablated versions of \IRMLM testing the three main components of \IRMLM: splitting the data into environments with one head per environment (\MTLM over \ERMLM), ensembling the heads during training (\EnsLM over \MTLM), using the specific competitive gradient update schedule (\IRMLM over \EnsLM). 
The four variants were run over all experiments previously described varying hyper-parameters yielding a total of $1320$ experimental results (see \Appref{app:details} for details) per architecture. To get a global view, we again aggregated these results with the paired aggregation given by the Bradley--Terry model. It estimates a strength for each architecture based on how likely it is to beat other architecture on the same experiments with the same hyper-parameters. It provides a scale-independent metric-independent way to aggregate scores \cite{peyrard-etal-2021-better} across tasks and experiments.
\begin{table}
\centering
\setlength{\tabcolsep}{3pt}
\begin{tabular}{@{}lccc|c@{}}
\toprule
        & \ERMLM & \MTLM  & \EnsLM  & \IRMLM\\ 
\midrule
\midrule
\ERMLM  & - & .92{\scriptsize± .06} & .26{\scriptsize± .09} & .28{\scriptsize± .09} \\
\MTLM  & .08{\scriptsize± .06} & - & .04{\scriptsize± .04} & .03{\scriptsize± .04} \\
\EnsLM  & .72{\scriptsize± .09} & .96{\scriptsize± .04} & - & .37{\scriptsize± .10} \\
\midrule
\IRMLM  & \textbf{.74{\scriptsize± .09}} & \textbf{.97{\scriptsize± .04}} & \textbf{.63{\scriptsize± .10}} & - \\
\bottomrule
\end{tabular}
\caption{\textmd{\textbf{Paired aggregated results.} Estimated probability that one architecture (row $i$) is better than any other (column $j$) across all previous experiments, based on the pairwise Bradley--Terry aggregation model.}}
\label{tab:bt-comparison}
\end{table}

The results are reported in \Tabref{tab:bt-comparison} and confirm the intuition built-up with previous experiments that simply having $n$ environments with $n$ heads is not beneficial on its own, as \MTLM does not provide benefits over \ERMLM. However, when combined with head ensembling (\EnsLM), significant improvements can be observed over both \ERMLM and \MTLM. Further significant benefits arise from the competitive gradient update specific to \IRMLM. While both \MTLM and \EnsLM have slightly better capacity to overfit with their $n$ heads, they don't benefit from the invariance regularization provided by competitive gradient updates.
Notice that \IRMLM is significantly better than any other architecture, as shown by the last row of \Tabref{tab:bt-comparison} (or equivalently, the last column).

\section{Discussion}
\label{sec:discussion}
In this section, we discuss our contributions in the context of previous work.

\subsection{Related Work}

\xhdr{Domain generalization}
The performance of deep learning models substantially degrades on out-of-domain (OoD) datasets, even in the face of small variations of the data\hyp generating process \cite{hendrycks2019benchmarking}.
\newcite{blanchard2011generalizing} have proposed domain generalization (DG) as a formalism for studying this problem. In DG, the goal is to learn a model using data from a single or multiple related but distinct training domains, in such a way that the model generalizes well to any OoD testing domain, unknown during training.
Recently, the problem of DG has attracted a lot of attention, and has been approached from different facets. 
Most of the existing methods fall under the paradigm of domain alignment \cite{inv_feature_rep, LiGTLT18, adv_inv_features, liu2020causal_semantic_rep, zhao2020entropy_reg}. Motivated by the idea that features that are stable across the training domains should also be robust to the unseen testing domains, these methods try to learn domain-invariant representations.
A group of other methods is based on meta-learning \cite{dou2019sem_features, balaji2018metareg, li2018learing_to_generalize}. 
The motivation behind this approach is that it exposes the model to domain shifts during training, which will allow it to generalize better during testing.
Regularization through data augmentation is commonly used in the training of machine learning models to alleviate overfitting and thereby improve generalization \cite{zhou2021feature_noise, zhou2020aug}. 
Based on this idea, \cite{zhou2021feature_noise, zhou2020aug} apply transformations on the original data to simulate a domain shift in training.

\xhdr{Domain generalization applied to language models}
In NLP, the default pipeline involves pretraining a task-agnostic language model, which is then finetuned on downstream tasks. This pretraining/finetuning division of learning is already known to improve robustness on downstream tasks \cite{hendrycks2019benchmarking}. However, the language models themselves suffer from spurious correlations and poor generalization even with small perturbations of the inputs \cite{moradi2021evaluating}.
To alleviate such problems, \newcite{oren-etal-2019-distributionally} adapted Distribution Robust Optimization \cite{dro} to language models. This resulted in a new loss minimizing the worst-case performance over subsamples of the training set. They focused on domains with topic shifts. Later, \newcite{vernikos-etal-2020-domain} used domain \hyp adversarial regularization to improve testing performance on unseen domains. 

Compared to these previous works, \IRMLM enjoys theoretical justification rooted in the causal framework of invariance \cite{icc}. Our implementation is simple, comes at negligible computational cost and can be applied directly to any transformer LM.

\subsection{Environment Design}
One question that might arise from the \IRMLM training schedule is what happens when environments have no lexical overlap. Maybe no correlation would remain for \IRMLM to model? 
We emphasize that \IRMLM learns a latent representation $\phi$ and stable correlations are those connecting this latent representation to observables, and not surface correlations between observables. To demonstrate that \IRMLM operates on latent variables and not just on surface-level correlations, we performed a simple experiment with languages as environments. We trained \IRMLM with a pretrained multilingual model (XLM-\ROBERTA) using English Wikipedia articles and Farsi Wikipedia articles as two environments. Despite almost no surface-level overlap, \IRMLM is still able to improve perplexity in each language individually and does not destroy previously learned correlations. This experiment is detailed in \Appref{app:en_fa}.

Also, if the number of environments grows arbitrarily large, certainly \IRMLM would not find any stable correlations in the data? We emphasize that the choice of environments is not intended to be arbitrary; simply contriving as many environments as possible could not be expected to be useful. Rather, the choice of environments has to reflect assumptions about the underlying data generation mechanism; \IRMLM then leverages the assumptions encoded in the choice of environments. 

Indeed, after this work has shown that \IRMLM can effectively remove unstable correlations, the next question becomes that of \textbf{environment design:} \emph{how to choose environment splits to be useful in practice?} 
Useful environment splits will likely be different for different tasks and different purposes. 
This work already demonstrated that the new paradigm of (i) environment design then (ii) \IRMLM is practical for language-related problems. 
Choosing environment splits is a flexible way to inject priors and inductive biases, compared to manually deciding which correlations are desired (as in bias removal) or fully learning the causal graph (as in causal reasoning).
Now, \IRMLM provides a computationally efficient framework to inject such priors and move the discussion from model inductive biases to data inductive biases. It already offers robustness to noise, a ready-to-use bias removal strategy for any existing language model needing few data points, and improves OoD generalization.

\section{Limitations}
\label{sec:limitations}
In this work, we focus on crafting controlled experiments with easily manageable dataset and language model sizes to carefully test the invariance benefits of \IRMLM. However, it is possible to expect different qualitative behavior for large-scale language models recently deployed due to emergent properties. 

Our implementation could be applied largely to various downstream tasks, other than language modeling measured by perplexity. 
Here, we focus on the language modeling task and perplexity measure because they allow clear and precise experiments measuring the ability of \IRMLM to deal with spurious correlations. The strong positive results observed in this work motivate future work to test \IRMLM in other setups closer to direct practical use-cases.

It is expected that different choices of environment splits will be useful for different downstream tasks. While this work demonstrates that \IRMLM is useful to remove spurious correlation, it does not say how to choose environments for which tasks. For instance, we observed smaller improvements when using thePile datasets and evaluating on the downstream GLUE tasks, indicating that thePile environment splits are not optimal for these downstream tasks. We believe that environment design is an important avenue for future research.


\section*{Acknowledgments}
This project is a collaboration between EPFL and Microsoft as part of the Microsoft Turing Academic Program (MS-TAP).
With support from
Swiss National Science Foundation (grant 200021\_185043),
European Union (TAILOR, grant 952215),
and gifts from Microsoft, Facebook, Google.

\bibliography{references}
\bibliographystyle{acl_natbib}

\appendix
\clearpage

\section{Illustration of \IRMLM Architecture}
\label{app:illustration}

\begin{figure*}[!t]
    \centering
    \includegraphics[width=0.99\linewidth]{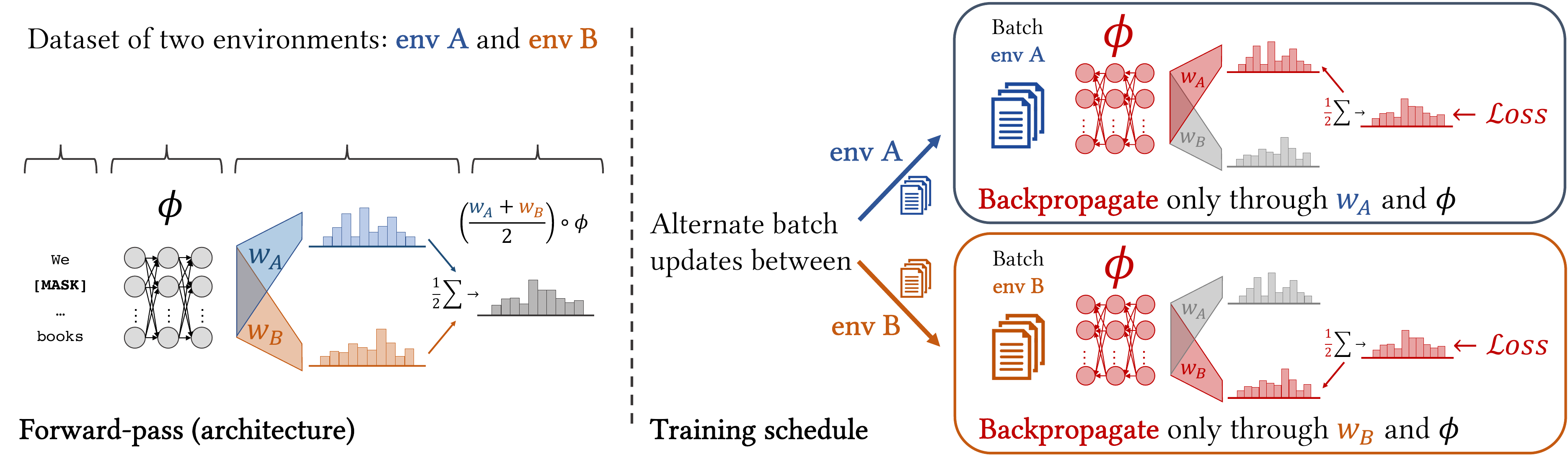}
    \caption{\textbf{Model description.} In the forward pass, input text goes through the main body of language model noted $\phi$ (e.g., a Transformer \cite{devlin-etal-2019-bert}), then one head per environment predicts logits over the vocabulary. These predictions are averaged over all heads and go through a softmax. During training, the model receives a batch of data from one environment $e$ and performs a gradient update only on the parameters of the main body of the language model ($\phi$) and on the parameters of the head tied to this environment $w_e$. Then batches are taken from each environment in a round-robin fashion.}
    \label{fig:model_desc}
\end{figure*}

In the main paper, we described formally the pseudo-code involved in training \IRMLM models. 
The model architecture and the logic of the training schedule is illustrated in \Figref{fig:model_desc} for the special-case of 2 environments ($n=2$).

\subsection{\MTLM and \EnsLM baselines}
We implemented two similar architectures that do not enjoy the same theoretical justifications. 

In the \MTLM baseline, the data is also split into $n$ environments with one head per environment. As in \IRMLM, environments take turns to send a batch of data and perform a batch update on the body of the transformer $\phi$ and the head associated with this environment. This is like viewing different environments as different tasks with uniform weights, even though they are all language modeling loss.

In the \EnsLM baseline, the data is split into $n$ environments with one head per environment. However, here, the heads are always predicting as an ensemble like \IRMLM. Here also the environments take turns to send a batch of data. The forward pass is exactly the same as the one of \IRMLM. In the backward pass, every head and the transformer body $\phi$ are always updated for every batch of every environment.

\section{Details about Experiments}
\label{app:details}

\subsection{Robustness to Noise}
\label{app:struct_noise}

\begin{figure*}
    \centering
    \includegraphics[width=0.85\linewidth]{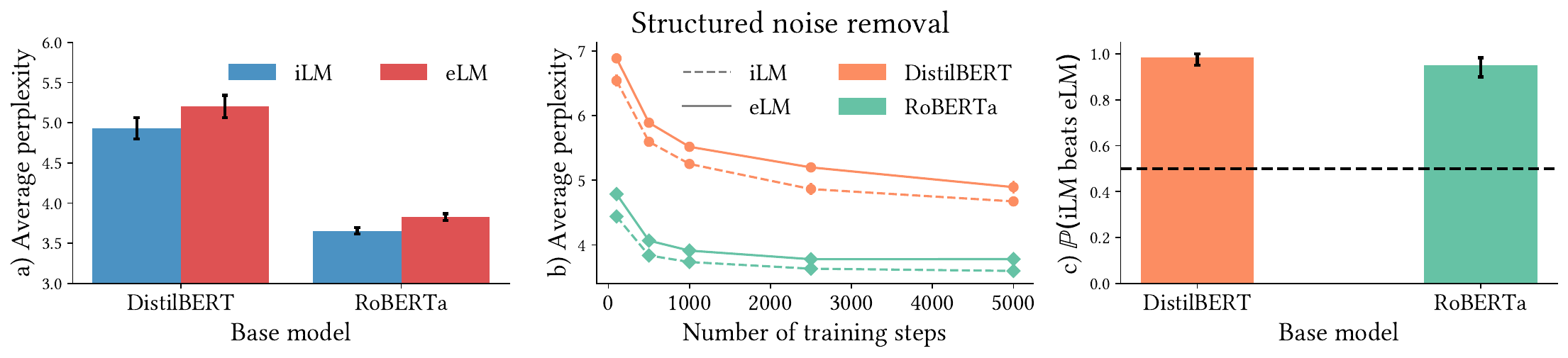}
    \caption{Structured noise removal experiment with environments having the same number of lines: a) average perplexity over all hyper-parameters b) average perplexity as a function of the number of training steps (for learning rate $10^{-5}$), c) Probability that \IRMLM is better than \ERMLM when compared on the same hyper-parameters}
    \label{fig:struct_noise_app}
\end{figure*}

\xhdr{Data}
The data used for this experiment comes from an HTML Wikipedia Dump of August 2018.
The files were pre-processed to remove the HTML content resulting in clean text articles. 
We randomly selected $60K$ articles with HTML (Env-B), and $60K$ different articles without HTML (Env-A). The test set contains $25K$ sentences coming from Wikipedia without HTML.

\xhdr{Hyper-parameters}
We ran the experiments reported in the main paper while varying several hyper-parameters: base transformers ($\phi$): [\dBERT, \ROBERTA{}], learning rates: $[1e^{-5}, 5e^{-5}]$, number of training steps: $[10, 100, 200, 500, 2500, 5000]$, $5$ random restarts with different random seeds, $2 \cdot 2 \cdot 6 \cdot 5 = 120$, ran with \ERMLM, \MTLM, \EnsLM, and \IRMLM resulting in $480$ experiments.

\xhdr{Number of lines vs. number of articles}
In the main paper, we report the results of \IRMLM and \ERMLM when trained with environments having the same number of articles.
However, the HTML articles have more lines and thus more \emph{sentences}.
Therefore, we also report in \Figref{fig:struct_noise_app} the same analysis repeated when the number of lines between Env-A and Env-B is the same, meaning Env-B contains fewer articles.
The conclusion remains largely unchanged in this scenario. As seen in \Figref{fig:struct_noise_app}~(c), \IRMLM has still a probability of beating \ERMLM for match hyper-parameters close to 1, highly significantly far away from $0.5$.

\subsection{Bias Removal}
\label{app:gb}

\xhdr{Data}
The dataset used for this experiment is Wikitext-2 \cite{MerityXBS16} where we follow the existing train/dev/test split. The dictionary of gendered terms comes from \newcite{bordia-bowman-2019-identifying} which was originally constructed to measure gender bias in language models.

The dictionary contains basic gender-pairs augmented with their variations in terms of casing, plural vs. singular forms and different spellings.
The basic gendered pairs are:
(actor, actress), (boy, girl), (boyfriend, girlfriend), (father, mother), (gentleman, lady), (grandson, granddaughter), (he, she), (hero, heroine), (him, her), (husband, wife),  (king, queen), (male, female), (man, woman), (mr., mrs.), (prince, princess), (son, daughter), (spokesman, spokeswoman), (stepfather, stepmother), (uncle, aunt)

\begin{table}
\centering
\setlength{\tabcolsep}{3pt}
\begin{tabular}{@{}lcccc@{}}
\toprule
        & 25 & 50  & 75  & 100\\ 
\midrule
\midrule
\multicolumn{3}{l}{\textbf{\dBERT}} \\ 
\hspace{4mm} \ERMLM  & .372{\scriptsize± .012} & .358{\scriptsize± .033} & .326{\scriptsize± .001} & \textbf{.308{\scriptsize± .016}} \\
\hspace{4mm} \MTLM  & .363{\scriptsize± .010} & .352{\scriptsize± .037} & .308{\scriptsize± .022} & .328{\scriptsize± .022} \\
\hspace{4mm} \EnsLM  & .322{\scriptsize± .003} & .350{\scriptsize± .032} & .324{\scriptsize± .020} & .315{\scriptsize± .015} \\
\hspace{4mm} \IRMLM  & \textbf{.309{\scriptsize± .006}} & \textbf{.322{\scriptsize± .033}} & \textbf{.318{\scriptsize± .012}} & .309{\scriptsize± .004} \\
\midrule
\multicolumn{3}{l}{\textbf{\ROBERTA}} \\ 
\hspace{4mm} \ERMLM  & .317{\scriptsize± .010} & .305{\scriptsize± .008} & .273{\scriptsize± .045} & \textbf{.259{\scriptsize± .025}} \\
\hspace{4mm} \MTLM  & .308{\scriptsize± .011} & .299{\scriptsize± .009} & .271{\scriptsize± .29} & .260{\scriptsize± .12} \\
\hspace{4mm} \EnsLM  & .291{\scriptsize± .011} & .300{\scriptsize± .011} & \textbf{.270{\scriptsize± .031}} & .271{\scriptsize± .033} \\
\hspace{4mm} \IRMLM  & \textbf{.290{\scriptsize± .013}} & \textbf{.291{\scriptsize± .003}} & .271{\scriptsize± .033} & .267{\scriptsize± .025} \\
\bottomrule
\end{tabular}
\caption{\textmd{\textbf{Complementary gender-bias removal results.} Average bias $B_H$ as described in \Secref{ssec:controlled_corr} across 4 different relative sizes of environments (25\%, 50\%, 75\% and 100\%).}}
\label{tab:gb_further}
\end{table}

\xhdr{Hyper-parameters}
We ran the experiments reported in the main paper while varying several hyper-parameters: base-model ($\phi$): [\dBERT, \ROBERTA{}], learning-rates: $[1e^{-5}, 5e^{-5}]$, number of training steps: $[10, 50, 100, 200, 1000, 2500]$, $5$ random restarts with different random seeds. This gives
$2 \cdot 2 \cdot 6 \cdot 5 = 120$ experimental parameters, ran for \ERMLM, \IRMLM, \MTLM, and \EnsLM while varying the relative sizes of environments in $[10, 25, 30, 50, 70, 75, 90, 100]$ resulting in $3840$ experiments.

\xhdr{Results for \MTLM and \EnsLM}
In \Figref{tab:gb_further}, we report the average bias as a function of the relative sizes of environments for \MTLM and \EnsLM alongside those of \IRMLM and \ERMLM.
We also observe here that \IRMLM outperform other architectures. Interestingly, \EnsLM seems to bring benefits in comparison to \ERMLM and \MTLM.

\begin{figure*}
    \centering
    \includegraphics[width=0.85\linewidth]{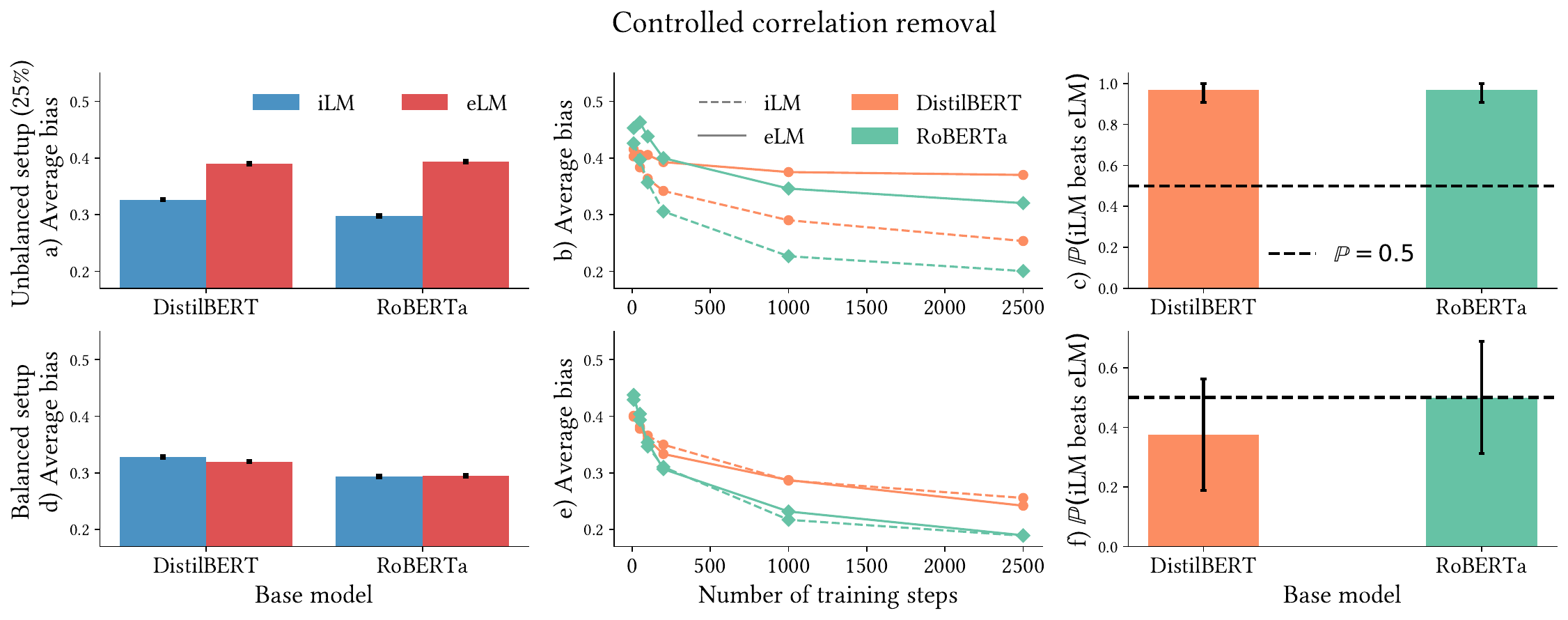}
    \caption{\textbf{Controlled correlation removal experiment}: 
    On the first row, the modified environment is 25\% of the size of the unmodified environment. On the second row, both have the same number of samples.
    On the left-most column, average bias over all hyper-parameters. 
    On the center column: average bias as a function of the number of training steps.
    On the right-most column: Probability that \IRMLM is less biased than \ERMLM when compared on the same hyper-parameters.
    }
    \label{fig:controlled_correlation_app}
\end{figure*}

\xhdr{Details about the results}
Here, we report complementary analysis compared to the results described in the paper. 
We report the performance of \ERMLM and \IRMLM as a function of the number of training steps and the probability that \IRMLM is better then \ERMLM when matched on hyper-parameter configuration as computed by the Bradley-Terry model.
This is reported by \Figref{fig:controlled_correlation_app} for two relative size: $25\%$ (the modified environment has $4$ times fewer examples) and $100\%$.

\xhdr{Perplexities after training}
To ensure that the gender-based correlations were not removed at the cost of a worse perplexity, we report in \Tabref{tab:gb_perplexities} the perplexities of \IRMLM models in comparison \ERMLM ones on the test set of Wikitext-2.
For reference, before our training \dBERT and \ROBERTA had, this same test set, perplexities of 14.25 and 6.92, respectively.

In \Tabref{tab:gb_perplexities}, the 95\% confidence intervals all give uncertainties $\approx 0.15$, meaning that for a fixed base model (\dBERT or \ROBERTA) all perplexities are within each other's error bounds. There is no significant perplexity difference between \ERMLM and \IRMLM or between the unbalanced and balanced setups.

\begin{table}
        \small
        \centering
        \begin{tabular}{l|c|c}
        \toprule
         & Unbalanced & Balanced  \\
        \midrule
        \IRMLM \ROBERTA & 4.16 & 4.13 \\
        \IRMLM \dBERT & 5.82 & 5.81 \\
        \ERMLM \ROBERTA & 4.14 & 4.14 \\
        \ERMLM \dBERT & 5.82 & 5.85 \\
        \bottomrule
        \end{tabular}
        \caption{Perplexities of \IRMLM and \ERMLM models after training.}
        \label{tab:gb_perplexities}
\end{table}

\subsection{Out-of-domain Generalization}
\label{app:ood}

\xhdr{Data}
The data used for this experiment comes from subsamples of thePile \cite{gao2020pile}. 

We randomly selected train and test domains as follow:
\begin{itemize}
    \item \textbf{Train}: "europarl", "freelaw", "dm mathematics", "youtubesubtitles", "USPTO backgrounds",
                        "arxiv", "books3", "wikipedia(en)", "stackexchange", "hackernews", "pile-cc"
    \item \textbf{Test}: "github", "ubuntu irc", "openwebtext2", "pubmed central",
                                        "enron emails", "pubmed abstracts", "gutenberg pg-19"
\end{itemize}

\xhdr{Hyper-parameters}
We ran the experiments reported in the main paper while varying several hyper-parameters: base-model ($\phi$): [\dBERT, \ROBERTA{}], learning-rates: $[1e^{-5}, 5e^{-5}]$, number of training steps: $[2500, 5000, 25000, 50000]$,  $5$ random restarts with different random seeds, for \ERMLM, \MTLM, \EnsLM, and \IRMLM. This results in $2\cdot2\cdot4\cdot5\cdot4=320$ experimental models, each evaluated in $3$ tasks: in-domain language modeling, out-of-domain language modeling, GLUE. This is a total of $960$ experimental setups.

\xhdr{Evaluation}
For the in-domain language modeling evaluation, we measure perplexity on 10K held-out sentences from each of the train domain.
Similarly for out-of-domain language modeling evaluation, we measure perplexity on 10K sentences from each of the test domain.

For GLUE, we used the default scripts from huggingface to evaluate trained models from checkpoints.

\subsection{Languages as Environments}
\label{app:en_fa}
One question that might arise from \IRMLM training schedule is whether it simply focuses on surface-level lexical correlations in the data. For example, if the lexical correlations are different across environments, maybe no correlation remain generalizable and \IRMLM learns an empty set of correlations. 
To better demonstrate that \IRMLM operate on latent variable and not on surface-level correlations, we perform a simple experiment with languages as environments.

\xhdr{Description}
We use two languages with no lexical overlap: English and Farsi. We put english Wikipedia articles as one environment and farsi Wikipedia articles as the other. In this setup, no surface-level correlations can generalize across environment as the two environments don't even have the same vocabulary.

We train \IRMLM with a multilingual pre-trained \ROBERTA: XLM-\ROBERTA for $5000$ steps with these two environments of equal size (10K articles per language). 
Then, we test whether this choice of environments destructs previously learn correlations in the language model by comparing perplexities on a balanced held-out test set of english and farsi documents against the model before finetuning. If the perplexities decrease, we would conclude that \IRMLM destroy surface-level correlations.

\xhdr{Results}
We found that before finetuning, XLM-\ROBERTA had a perplexity of $14.56$ on the held-out test set, where \IRMLM could improve it perplexity down to $6.44$. This indicates that \IRMLM with environments having no lexical overlap does not destroy previously learned correlations. It can even improve its perplexities for each language.
A possible reason why \IRMLM can even improve so dramatically compared to before finetuning might come from the fact that $\phi$ learns to recognize the languages, separate them and treat them separately. Similar effects have been observed in previous work \cite{OodIRM} when the correlation between the environment index and the target variable is very strong (which is the case here).

\subsection{Head dynamics}
\label{app:heads_drifting}

\begin{figure}
    \centering
    \includegraphics[width=0.80\linewidth]{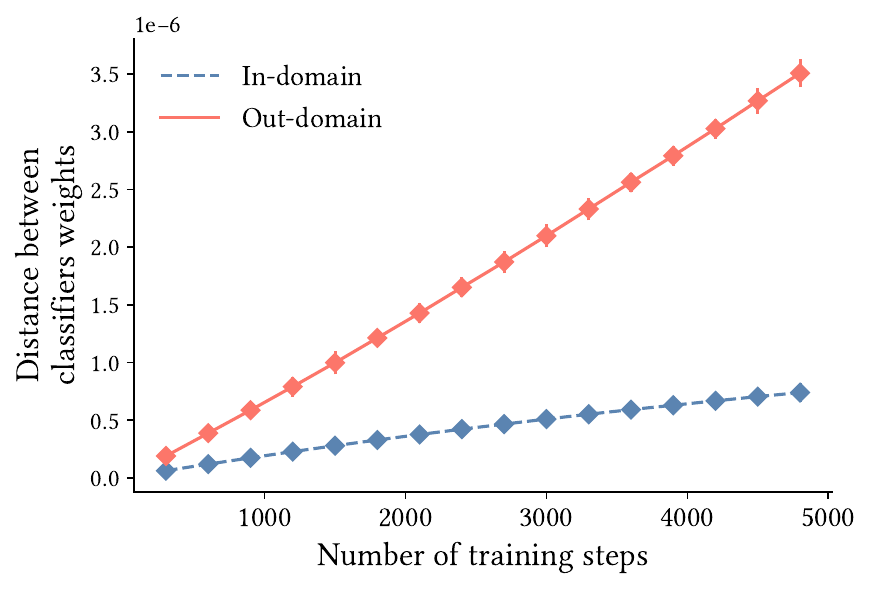}
    \caption{Comparing distance between heads weights in- and out-domain as functions of the number of training step. (95\% confidence interval from random restart with different seeds.)}
    \label{fig:heads_drifting}
\end{figure}

The main components of our framework are the heads and their training dynamic. Therefore, we investigate aspects related to behaviour of the heads. 

\xhdr{Description}
During training, the loss of each head is still entangled with the prediction of every other head. So we wonder whether the heads still capture information related to the environment it is tied to during training.
In particular, we ask 
(i) whether the parameters of the heads for different environments are drifting apart during training? Indeed, all heads are initialized to the same pretrained weights at the beginning of training.
(ii) Are the parameters of the heads predicting which environments are more similar?

\xhdr{Experimental setup}
To answer these two questions in one go, we take two environments $A$ and $B$ and split each of them into two new environments resulting in $A_1$, $A_2$, $B_1$, and $B_2$ such that $A_1$ and $A_2$ are very similar $B_1$ and $B_2$ are very similar but $A_i$ and $B_i$ are different. 
We then train \IRMLM with the four environments and, thus, with four heads $w_{A_1}$, $w_{A_2}$, $w_{B_1}$, and $w_{B_2}$. 
We measure whether the heads’ weights can predict the similarities between A’s and B’s environments.
\begin{align}
    &D_{in} = \frac{1}{2} \left(d(w_{A_1}, w_{A_2}) + d(w_{B_1}, w_{B_2}) \right), \\
    &D_{out} = \frac{1}{4} \sum\limits_{i, j} d(w_{A_i}, w_{B_j}),
\end{align}
where $d$ is the L2 distance between the linearized weights of two heads.
Then, $D_{in}$ is the average distance between heads tied the same domain, and $D_{out}$ is the average distance between heads tied to different domains. Remember that in this case, there are 2 domains $A$ and $B$ and 4 environments $A_i$ and $B_i$.

In this experiment, we randomly select the base environments $A$ and $B$ from the domains of thePile ($A$ is the Enron-Email, and $B$ is PubMed abstract). We create $A_i$ and $B_i$ by randomly subsampling $2$ environments of the same size from each domain.
We train \IRMLM with \ROBERTA{} for $5000$ training steps, taking checkpoints of the heads every $500$ steps. We perform $10$ random restarts with different seeds to uncertainty estimates.
In \Figref{fig:heads_drifting}, we report $D_{in}$ and $D_{out}$ as functions of the number of training steps. 

\begin{figure}[!htb]
    \centering
    \includegraphics[width=0.90\linewidth]{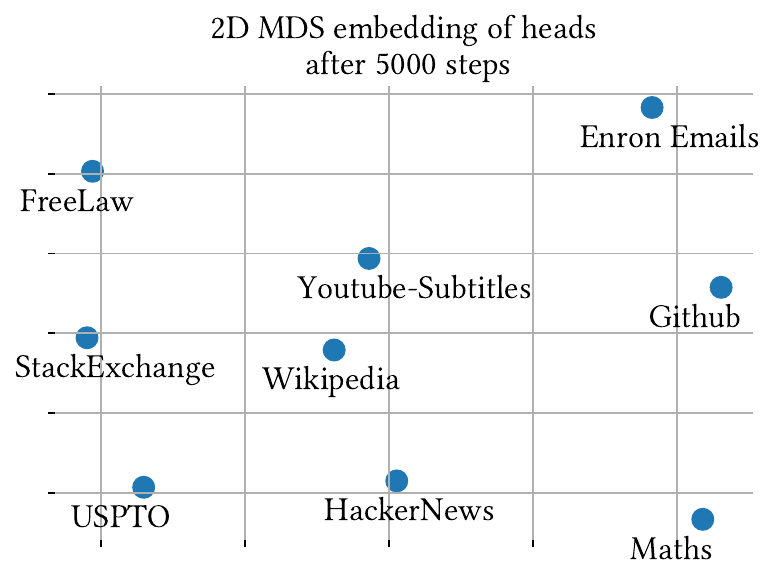}
    \caption{Heads embeddings: 2D projection of the heads parameters similarity structure after training \IRMLM with \ROBERTA for $5000$ steps with $9$ domains. Each dot represent one head of the model after training and the labels indicate to which domain it is tied to.}
    \label{fig:head_mds}
\end{figure}

\xhdr{Analysis}
We first notice that indeed the heads are drifting apart from each other as training advances. 
More interestingly, the distance between heads from the same domain is significantly much smaller than the distance between heads from different domains. We conclude that heads retain environment-specific information in their parameters and are predictive of environment similarities.

Now, we visualize the geometry of head similarity by training \IRMLM with \ROBERTA for $5000$ steps with $9$ environments from thePile: .
After training, we take the heads' parameters and compute the pairwise distance between all $9$ heads and embed them in 2D with Multi-Dimensional Scaling to visualize the similarity structure. The result is depicted in \Figref{fig:head_mds}.
\FloatBarrier


\end{document}